\documentclass [10pt,twocolumn,twoside]{IEEEtran}
\usepackage{orcidlink}
\usepackage[numbers,sort&compress]{natbib}
\usepackage[numbers]{natbib}
\usepackage{amssymb}
\usepackage{mathrsfs}
\usepackage{graphics}
\usepackage{graphicx}
\usepackage{subfigure}
\usepackage{epsfig}
\usepackage{epstopdf}
\usepackage{amsmath}
\usepackage{algorithmic}
\usepackage{hyperref}
\usepackage{algorithm}
\usepackage{bm}
\usepackage{lineno}
\usepackage{url}
\usepackage{float}
\usepackage{color}
\usepackage{amssymb}
\usepackage{multirow}
\usepackage{subfigure}
\usepackage{graphicx}
\usepackage{amsmath}
\usepackage{url}
\usepackage{caption}
\usepackage[normalem]{ulem}
\usepackage[color,final]{showkeys}
\usepackage{multirow}
\usepackage{setspace}
\ifCLASSINFOpdf
\else
\fi
\begin{document}
\title{Matrix Completion Via Reweighted Logarithmic Norm Minimization}

\author{Zhijie Wang\orcidlink{https://orcid.org/0009-0004-4066-962X},  ~\IEEEmembership{}   
Liangtian He\orcidlink{https://orcid.org/0000-0002-1300-1892},  ~\IEEEmembership{}  
Qinghua Zhang\orcidlink{https://orcid.org/0009-0003-9647-2300},  ~\IEEEmembership{}    
Jifei Miao\orcidlink{https://orcid.org/0000-0001-5663-8749},  ~\IEEEmembership{}   Liang-Jian Deng\orcidlink{https://orcid.org/0000-0003-3178-9772},  ~\IEEEmembership{}   
Jun Liu\orcidlink{https://orcid.org/0000-0003-2073-8320}  ~\IEEEmembership{}  
%
%
%
%
%
}

\markboth{IEEE Signal Processing Letters}%
{Wang \MakeLowercase{\textit{et al.}}: Matrix Completion Via Reweighted Logarithmic Norm Minimization}

\maketitle

\begin{abstract}
Low-rank matrix completion (LRMC) has demonstrated remarkable success in a wide range of applications. 
To address the NP-hard nature of the rank minimization problem, the nuclear norm is commonly used as a convex and computationally tractable surrogate for the rank function. However, this approach often yields suboptimal solutions due to the excessive shrinkage of singular values. In this letter, we propose a novel reweighted logarithmic norm as a more effective nonconvex surrogate, which provides a closer approximation than many existing alternatives. We efficiently solve the resulting optimization problem by employing the alternating direction method of multipliers (ADMM). Experimental results on image inpainting demonstrate that the proposed method achieves superior performance compared to state-of-the-art LRMC approaches, both in terms of visual quality and quantitative metrics.
\end{abstract}
\section{Introduction}
\IEEEPARstart{R}{covering}  the missing entries of a matrix from incomplete observations, known as matrix completion, has garnered considerable attention in various applications  \cite{ramlatchan2018survey,chi2018low,jiang2017robust,li2020rank,fathi2021two}. 
Among these, low-rank matrix completion (LRMC) has demonstrated to be an effective technique. Current LRMC methods can be broadly categorized into two types. The first type employs the matrix factorization technique \cite{Shang2018DN, Chen2022RLMF,Wen2012LF,Ma2014Decomposition,Chi2019LF}, which decomposes a matrix into the product of two or more smaller matrices to enhance computational efficiency. 
However, this approach typically requires pre-specifying the underlying rank, a task that can be challenging in practice.

The second type of LRMC methods is based on spectral regularization \cite{Lu2014,li2020matrix,Nie2012,Nie2018}. 
To avoid the NP-hard rank minimization problem, the nuclear norm is commonly adopted as a convex surrogate, leading to a computationally tractable nuclear norm minimization (NNM) problem \cite{Cai2010SVT}. 
However, the NNM applies uniform shrinkage to all singular values, which may lead to suboptimal solutions. 
Consequently, to more accurately approximate the rank function, a variety of nonconvex surrogate functions have been proposed in the literature,
including the truncated nuclear norm \cite{Hu2013TNNM},  the weighted nuclear norm \cite{Gu2017WNNM},  the log-determinant heuristic \cite{yang2019weakly,chen2014convergence,kang2016top}, the Schatten capped
$p$-norm  \cite{li2020matrix}, and the nuclear norm minus Frobenius norm \cite{shan2023multi}, to mention just a few. 
These functions, in principle, penalize larger singular values less and smaller singular values more, and have empirically demonstrated superior performance compared to NNM.

Very recently, Chen et al. \cite{chen2021logarithmic} proposed an efficient algorithm termed logarithmic norm regularized matrix factorization (LRMF), in which the authors
introduced an effective matrix logarithmic norm (MLN) and combined it with bi-factor and multi-factor strategies to enhance computational efficiency. 
However, the LRMF method is restricted to specific values of $p$, that is,  $p = \frac{1}{n}$, where $n = 1,2,\cdots$.

In this letter, to make the MLN surrogate proposed in \cite{chen2021logarithmic} a better approximation of the rank function, we propose a novel \textit{reweighted} matrix logarithmic norm (RMLN). Additionally, we develop an iterative optimization algorithm within the framework of the alternating direction method of multipliers (ADMM), which enables our proposed method to be applicable to any $0<p\leqslant 1$. 
The main contributions of this work are summarized as follows.
\begin{itemize}
\item We propose a novel reweighted matrix logarithmic norm regularization, which provides a more accurate approximation to the rank function.
    
\item The resulting optimization problem is solved within the ADMM framework. 
      In contrast to the LRMF algorithm proposed in \cite{chen2021logarithmic}, our approach is applicable to any  $p \in  (0,1]$, thereby removing the restriction to specific values.

\item Extensive experiments on image inpainting demonstrate the superior recovery performance of our proposed method compared to state-of-the-art LRMC approaches.
\end{itemize}

\section{Preliminaries}
\subsection{Notations}
Throughout this letter, $\mathbb{R}$ denotes the real space. 
Scalars, vectors, and matrices are denoted by $x$, $\mathbf{x}$, and $\mathbf{X}$, respectively. 
Standard operations include $(\cdot)^{-1}$ for the matrix inversion and $(\cdot)^{T}$ for the transposition.
The Frobenius norm and the nuclear norm are denoted by $||\cdot||_F$ and $||\cdot||_{\ast}$, respectively. Additionally, $\mathrm{det}(\cdot)$ represents the matrix determinant, and $tr(\cdot)$ denotes the matrix trace.

\subsection{LRMC Problem}
The LRMC problem aims to recover the underlying  matrix $\mathbf{X} \in \mathbb{R}^{M\times N}$ from its partially observed entries in $\mathbf{Y}$, indexed by the set $\Omega$. 
This problem can be formulated as:
\begin{equation}\label{eq: rank model}
\min_{\mathbf{X}}  \mathrm{rank}(\mathbf{X})  \quad \text{s.t.} \quad  \mathcal{P}_{\Omega}(\mathbf{X} - \mathbf{Y}) = \mathbf{0},
\end{equation}
where $\mathrm{rank}(\cdot)$ denotes the rank of a matrix. 
The projection operator $\mathcal{P}_{\Omega}: \mathbb{R}^{M\times N} \to \mathbb{R}^{M\times N}$ is defined as:
\begin{equation} \label{eq: projection operator}
(\mathcal{P}_{\Omega}(\mathbf{X}))_{i,j} = \left\{ \begin{array}{ll}
\mathbf{X}_{i,j}, & (i,j) \in \Omega,      \\
0, & (i,j) \in \Omega^{c}, \end{array} \right.
\end{equation} 
where $\mathbf{X}_{i,j}$ denotes the entry at position $(i,j)$,  and $\Omega^{c}$ represents the complement of the set $\Omega$.

However, the optimization problem specified in Eq. \eqref{eq: rank model} is NP-hard, owing to the non-convexity and discontinuity inherent in the rank function. 
To tackle this intractability, the nuclear norm is commonly employed as a convex surrogate. It is formally defined as the sum of singular values, i.e.,  $||\mathbf{X}||_{\ast} = \sum_{i=1}^{\min\{M,N\}} \sigma_{i}(\mathbf{X})$, where $\sigma_{i}(\mathbf{X})$ denotes the $i$-th singular value of $\mathbf{X}$.
A key justification for its use is that it constitutes the tightest convex envelope of the rank function.
This leads to the nuclear norm minimization (NNM) recovery formulation:
\begin{equation}\label{eq: NNM model}
\min_{\mathbf{X}}  ||\mathbf{X}||_{\ast}   \quad \text{s.t.} \quad  \mathcal{P}_{\Omega}(\mathbf{X} - \mathbf{Y}) = \mathbf{0}.
\end{equation}
While NNM renders the optimization problem computationally 
tractable \cite{Cai2010SVT, oh2017fast}, a notable drawback is its tendency to excessively shrink larger singular values. 
This bias often leads to suboptimal solutions and thus motivates the search for more precise nonconvex surrogates that can better approximate the rank function.

\subsection{Matrix Logarithmic Norm}
Motivated by the log-determinant heuristic $\mathrm{log}\mathrm{det}(\mathbf{X} + \varepsilon\mathbf{I})$ \cite{kang2016top}, Chen et al. \cite{chen2021logarithmic} introduced the matrix logarithmic norm (MLN) as a nonconvex surrogate for the rank function.

\textbf{Definition 1.} (MLN \cite{chen2021logarithmic}) 
\textit{Given a matrix $\mathbf{X} \in \mathbb{R}^{M\times N}$, the MLN of $\mathbf{X}$ is defined as:
\begin{equation}\label{eq: MLN} 
\small
||\mathbf{X}||^p_{L} := \sum_{i=1}^{\mathrm{min}\{M,N\}}\log(\sigma_{i}^p(\mathbf{X}) + \varepsilon),
\end{equation}
where $\sigma_{i}(\mathbf{X})$ denotes the $i$-th singular value of $\mathbf{X}$. }

To facilitate optimization, the authors derived both bi-factor and multi-factor matrix factorization forms for the MLN and subsequently employed a block-coordinate descent algorithm to efficiently solve the resulting minimization problem. 
A notable restriction of this approach, however, is that the parameter $p$ is restricted to specific discrete values, namely $p = \frac{1}{n}$ for $n = 1,2,\cdots$.
For comprehensive algorithmic details, the interested readers are referred to \cite{chen2021logarithmic}.

\section{The Proposed Method}
\subsection{Reweighted Matrix Logarithmic Norm}
Drawing inspiration from the success of reweighting strategies \cite{xie2016weighted,Chen2022RLMF,Huang2020reweighted} and the effective rank approximation provided by the MLN \cite{chen2021logarithmic}, we propose the \textit{reweighted} matrix logarithmic norm (RMLN). 
This new surrogate function is formally defined as follows.

\textbf{Definition 2.} (RMLN) 
\textit{Given a matrix $\mathbf{X} \in \mathbb{R}^{M\times N}$, the RMLN of $\mathbf{X}$ is defined as:
\begin{equation}\label{eq: RLN} 
\small
||\mathbf{X}||^p_{\mathbf{w}, L} := \sum_{i=1}^{\mathrm{min}\{M,N\}}w_{i}\log(\sigma_{i}^p(\mathbf{X}) + \varepsilon),
\end{equation}
where the weights are given by:
\begin{equation}\label{eq: reweights design}
w_{i} = \gamma\left(\log(\sigma_{i}^p(\mathbf{X}) + \varepsilon) + c \right)^{p-1},  \  0<p\leqslant1, 
\end{equation}
where $c$, $\gamma$, and $\varepsilon$ are positive constants. }

\begin{figure}
\centering
\includegraphics[width=0.2\textwidth]{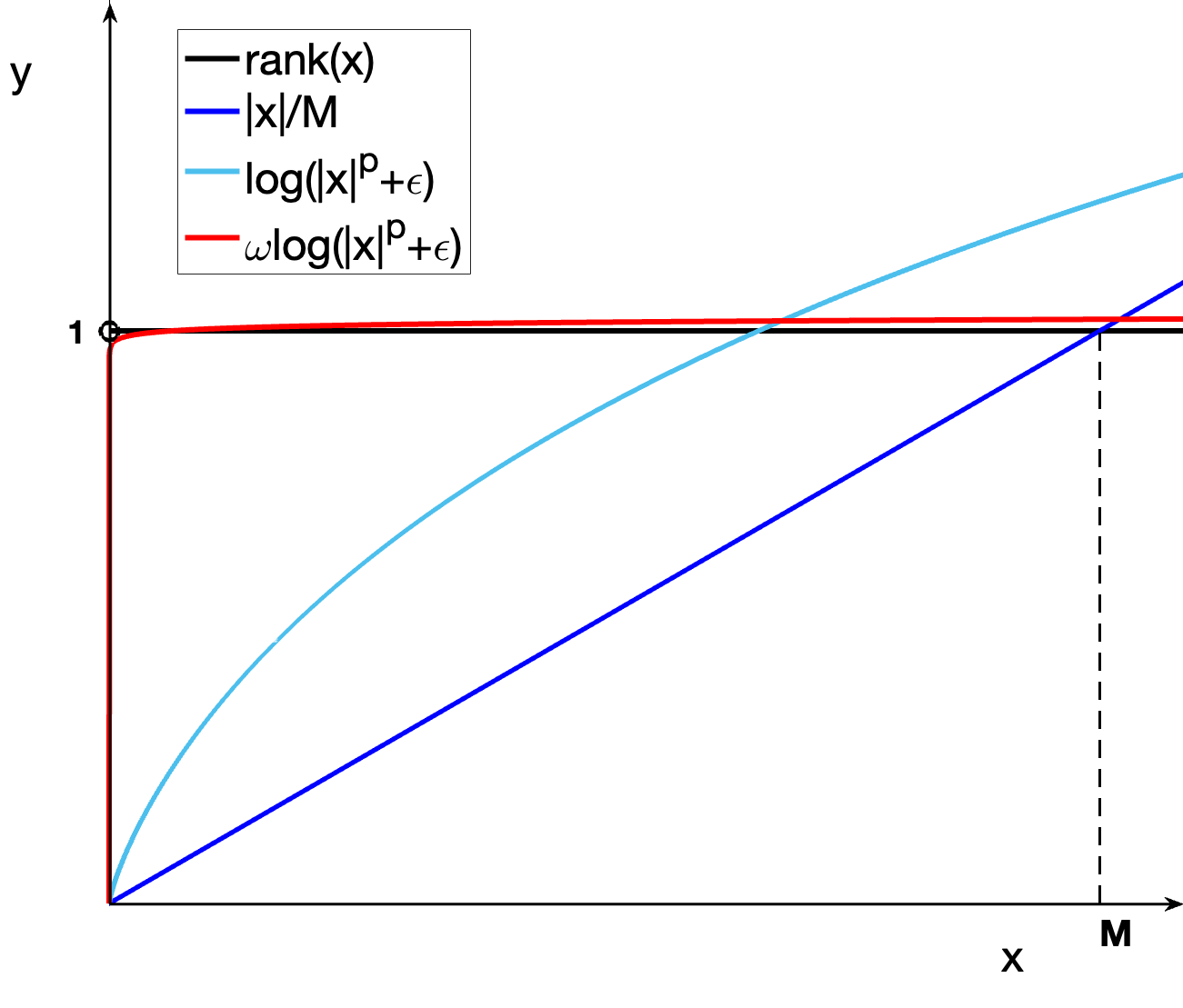}

\caption{\small Comparison of the rank function, the convex envelope of rank (nuclear norm) \cite{Cai2010SVT}, the MLN \cite{chen2021logarithmic}, and the proposed RMLN for scalar $x$.}\label{FG:rank.png}
\end{figure}

Fig. \ref{FG:rank.png} provides an intuitive comparison of the rank function, the nuclear norm, the MLN \cite{chen2021logarithmic}, and the proposed RMLN in the scalar case, where $\mathbf{X} = x \in \mathbb{R}$. 
For a scalar $x$ bounded by $M$ (i.e.,  $|x| \leqslant M$), then $\frac{||\mathbf{X}||_{\ast}}{M} = \frac{|x|}{M}$ forms the convex envelope of the rank function \cite{chen2021logarithmic,fazel2001rank}, and the MLN and RMLN are given by $||\mathbf{X}||_{L}^p = \log(|x|^p+\varepsilon)$ and $||\mathbf{X}||^p_{\mathbf{w}, L} = w\log(|x|^p+\varepsilon)$, respectively. 
As observed, our proposed RMLN exhibits a behavior that closely approximates the rank function, which highlights its superior capability for rank approximation.

\subsection{RMLN for Matrix Completion }
We employ the proposed RMLN as a spectral regularization and formulate the matrix completion problem as follows:
\begin{equation}\label{eq:RMLNMC}
\min_{\mathbf{X}}  \lambda||\mathbf{X}||^p_{\mathbf{w},L} + \frac{1}{2}  ||\mathcal{P}_{\Omega}(\mathbf{X} - \mathbf{Y})||^2_F.
\end{equation}
To solve this optimization problem, we adopt the ADMM \cite{Boyd2011ADMM}. 
Specifically, by introducing an auxiliary variable $\mathbf{Z}$, it allows us to reformulate the original problem (\ref{eq:RMLNMC}) into an equivalent constrained form:
\begin{equation}
\min_{\mathbf{X},\mathbf{Z}}  \lambda||\mathbf{Z}||^p_{\mathbf{w},L} + \frac{1}{2}  ||\mathcal{P}_{\Omega}(\mathbf{X} - \mathbf{Y})||_F^2, \quad s.t. \quad \mathbf{Z} = \mathbf{X}.
\end{equation}
The corresponding augmented Lagrangian for this constrained problem is given by:
\begin{equation}
\small
\mathcal{L}_{\mu}(\mathbf{X},\mathbf{Z},\mathbf{\Lambda}) = \frac{1}{2}  ||\mathcal{P}_{\Omega}(\mathbf{X} - \mathbf{Y})||_F^2 + \lambda||\mathbf{Z}||^p_{\mathbf{w},L} + \frac{\mu}{2}||\mathbf{Z} - \mathbf{X} - \frac{\mathbf{\Lambda}}{\mu} ||_F^2,
\end{equation}
where $\mu > 0$ is a penalty parameter and $\mathbf{\Lambda}$ is the Lagrangian multiplier. The ADMM algorithm then proceeds by iteratively minimizing $\mathcal{L}_{\mu}$ with respect to $\mathbf{X}$ and $\mathbf{Z}$ alternately, while updating the multiplier $\mathbf{\Lambda}$.

In the $k$-th iteration, with the other variables fixed, this procedure yields the following update rules.

\textit{1) Updating} $\mathbf{X}$: 
\begin{equation}\label{Xsub}
\small
	\mathbf{X}^{(k+1)} = \mathrm{arg} \min_{\mathbf{X} }\frac{1}{2}||\mathcal{P}_{\Omega}(\mathbf{X} - \mathbf{Y})||_{F}^{2} + \frac{\mu^{(k)}}{2}||\mathbf{X}+ \frac{\mathbf{\Lambda}^{(k)}}{\mu^{(k)}} - \mathbf{Z}^{(k)}||_F^{2}.
\end{equation}
Applying the first-order optimality condition yields a closed-form solution:
\begin{equation}\label{eq:Xrewrite}
\small
	\mathbf{X}^{(k+1)} = \mathcal{P}_{\Omega^{c}}\left(\mathbf{Z}^{(k)} - \frac{\mathbf{\Lambda}^{(k)}}{\mu^{(k)}}\right) + \mathcal{P}_{\Omega}\left(\frac{\mathbf{Y} + \mu^{(k)}\mathbf{Z}^{(k)} - \mathbf{\Lambda}^{(k)}}{1 + \mu^{(k)}}\right).
\end{equation}

\textit{2) Updating} $\mathbf{Z}$: 
\begin{equation}\label{Zsub}
\small
	\mathbf{Z}^{(k+1)} = \mathrm{arg} \min_{\mathbf{Z}} \frac{\mu^{(k)}}{2}||\mathbf{Z} - \mathbf{X}^{(k+1)} - \frac{\mathbf{\Lambda}^{(k)}}{\mu^{(k)}} ||_F^2 + \lambda||\mathbf{Z}||^p_{\mathbf{w},L}.
\end{equation}
Before deriving the solution to Eq. (\ref{Zsub}), we first present the following theorem.

\textbf{Theorem 1.} 
\textit{Let $\mathbf{Y} \in \mathbb{R}^{M \times N}$ be any matrix with singular value decomposition $\mathbf{Y} = \mathbf{U}_{\mathbf{Y}}\mathbf{\Sigma}_{\mathbf{Y}}\mathbf{V}_{\mathbf{Y}}^T$, where $\mathbf{\Sigma}_{\mathbf{Y}} = diag(\sigma_1(\mathbf{Y}), \sigma_2(\mathbf{Y}), \cdots, \sigma_{\mathrm{min}\{M,N\}}(\mathbf{Y}))$.  
For any $\eta > 0$, the optimal solution to the following optimization problem:
\begin{equation} \label{RMLNpro}
\small
	\mathrm{arg} \min_{\mathbf{X}} \frac{1}{2}||\mathbf{X} - \mathbf{Y}||_F^2 + \eta||\mathbf{X}||^p_{\mathbf{w},L}
\end{equation}
is given by $\mathbf{X}^{\ast} = \mathbf{U}_{\mathbf{Y}}\mathbf{\Sigma}_{\mathbf{X}}^{\ast}\mathbf{V}_{\mathbf{Y}}^T$. 
Here, the singular value matrix $\mathbf{\Sigma}_{\mathbf{X}}^{\ast} = diag(\sigma^{\ast}_1(\mathbf{X}),\sigma^{\ast}_2(\mathbf{X}),\cdots,\sigma^{\ast}_{\mathrm{min}\{M,N\}}(\mathbf{X}))$  and the optimal singular values $\sigma_i^{\ast}(\mathbf{X})$ are obtained by:
\begin{equation} \label{RMLNproeqi}
\small
	\sigma_i^{\ast}(\mathbf{X}) = \mathrm{arg} \min_{\sigma_i(\mathbf{X}) \geqslant 0} \frac{1}{2} (\sigma_i(\mathbf{X}) - \sigma_i(\mathbf{Y}))^2 + \eta w_{i}\log(\sigma_{i}^p(\mathbf{X}) + \varepsilon),
\end{equation}
where $i = 1,2,\cdots, \mathrm{min}\{M,N\}$. }

\textbf{\textit{Proof.}} 
\textit{
Let $\mathbf{Y} = \mathbf{U}_\mathbf{Y}\mathbf{\Sigma}_\mathbf{Y}\mathbf{V}^T_\mathbf{Y}$ be the SVD of $\mathbf{Y} \in \mathbb{R}^{M \times N}$. 
Then, we have:
\begin{flalign}
\small
&\begin{aligned}
	&\frac{1}{2}||\mathbf{X} - \mathbf{Y}||_F^2 + \eta||\mathbf{X}||^p_{\mathbf{w},L} \\
	& = \frac{1}{2}\left(tr(\mathbf{Y}^T\mathbf{Y}) + tr(\mathbf{X}^T\mathbf{X}) - 2tr(\mathbf{X}^T\mathbf{Y})\right) + \eta||\mathbf{X}||^p_{\mathbf{w},L} \\
	& = \frac{1}{2}\left(\sum_i\sigma^2_i(\mathbf{Y}) + \sum_i\sigma^2_i(\mathbf{X}) -2tr(\mathbf{X}^T\mathbf{Y})\right) + \eta||\mathbf{X}||^p_{\mathbf{w},L} \\
	& \geqslant \frac{1}{2}\sum_i\left(\sigma^2_i(\mathbf{Y}) + \sigma^2_i(\mathbf{X}) - 2\sigma_i(\mathbf{Y})\sigma_i(\mathbf{X})\right) + \eta||\mathbf{X}||^p_{\mathbf{w},L} \\
	& = \sum_i\frac{1}{2}(\sigma_i(\mathbf{Y}) - \sigma_i(\mathbf{X}))^2 + \eta w_{i}\log(\sigma_{i}^p(\mathbf{X}) + \varepsilon), \notag
\end{aligned}&
\end{flalign}
in which the inequality follows from Von Neumann's trace inequality \cite{mirsky1975trace}. Therefore,  the optimal solution to Eq. (\ref{RMLNpro}) is given by $\mathbf{X}^{\ast} = \mathbf{U}_{\mathbf{Y}}\mathbf{\Sigma}_{\mathbf{X}}^{\ast}\mathbf{V}_{\mathbf{Y}}^T$.  $\blacksquare$ }

According to Theorem 1, the problem in Eq. (\ref{RMLNpro}) is equivalent to solving Eq. (\ref{RMLNproeqi}). 
Note that the function $w\log(|x|^p + \varepsilon)$ is concave, monotonically increasing, and continuously differentiable on $[0, +\infty)$  
These properties render the problem Eq. (\ref{RMLNproeqi}) amenable to the difference of convex (DC) algorithm \cite{tao1997convex}.
Specifically, the DC algorithm proceeds by iteratively linearizing the concave part of the objective function, while leaving the convex quadratic $\ell_2$-norm term unchanged. 
This results in the following iterative update for each singular value:
\begin{equation}\label{eq:DC subproblem}
\small
	\sigma_i^{(t+1)}(\mathbf{X}) = \mathrm{arg} \min_{\sigma_i(\mathbf{X}) \geqslant 0} \frac{1}{2}(\sigma_i(\mathbf{X}) - \sigma_i(\mathbf{Y)})^2 +  \frac{\eta w_ip(\sigma_i^{(t)}(\mathbf{X}))^{p-1}\sigma_i}{(\sigma_i^{(t)}(\mathbf{X}))^{p} + \varepsilon} ,
\end{equation}
where $i = 1,2,\cdots, \min\{M,N\}$, and the superscript $t$ is the iteration index.

A closed-form solution for Eq. (\ref{eq:DC subproblem}) can be derived by applying the first-order optimality condition, leading to the final update rule:
\begin{equation}\label{eq:sigmasovling}
\small
	\sigma_i^{(t+1)}(\mathbf{X}) = \mathrm{max}\left\{\sigma_i(\mathbf{Y}) -  \frac{\eta w_i p(\sigma_i^{(t)}(\mathbf{X}))^{p-1}}{(\sigma_i^{(t)}(\mathbf{X}))^{p} + \varepsilon}, 0\right\}.
\end{equation}

\textit{3) Updating} $\mathbf{\Lambda}$ and $\mu$: 
\begin{equation}\label{eq: Lagupdate} 
\small
\left\{\begin{array}{ll}
\mathbf{\Lambda}^{(k+1)} = \mathbf{\Lambda}^{(k)} + \mu^{(k)}(\mathbf{X}^{(k+1)} - \mathbf{Z}^{(k+1)}),   \\
\mu^{(k+1)} = \mu^{(k)}\cdot\rho \quad(\rho>1).
\end{array}\right.
\end{equation}

For clarity, the complete procedure for the proposed RMLN regularization based matrix completion is detailed in Algorithm \ref{Alg:RLNM}.
 
\begin{algorithm}[]
\caption{RMLN minimization for matrix completion.}	
\label{Alg:RLNM}
\renewcommand{\algorithmicrequire}{\textbf{Input:}}
\renewcommand{\algorithmicensure}{\textbf{Output:}}
\begin{algorithmic}[1]
\REQUIRE Incomplete matrix $\mathbf{Y} \in \mathbb{R}^{M \times N}$, the set of observed entries $\Omega$, $\mu^{(0)}$, $\rho$, $\gamma$, $\lambda$, $p$, $\varepsilon$, $c$,  $K$ and $T$.
\STATE \textbf{Initialize:} $\mathbf{X}^{(0)}$ = $\mathbf{Z}^{(0)}$ = $\mathbf{Y}$, $\mathbf{\Lambda}^{(0)}$ = $\mathbf{0}$.
\FOR {$k=0,1,\cdots, K$}
\STATE Update $\mathbf{X}^{(k+1)}$ by Eq. (\ref{Xsub}).
\STATE Update $\mathbf{Z}^{(k+1)}$ by Eq. (\ref{Zsub}) based on Theorem 1.
\FOR {$i = 1,2,\cdots, \mathrm{min}\{M,N\}$}
\STATE Update $w_i = \gamma\left(\log(\sigma_{i}^p(\mathbf{Z}^{(k)}) + \varepsilon) + c \right)^{p-1}$.
\STATE $\sigma_i^{(0)}(\mathbf{Z}^{(k+1)}) = \sigma_i(\mathbf{Z}^{(k)})$.
\FOR {$t=0,1,\cdots,T$}
\STATE Update $\sigma_i^{(t+1)}(\mathbf{Z}^{(k+1)}) = \mathrm{max}\{(\sigma_i(\mathbf{X}^{(k+1)} + \frac{\mathbf{\Lambda}^{(k)}}{\mu^{(k)}}) -  \frac{\lambda w_i p(\sigma_i^{(t)}(\mathbf{Z}^{(k+1)}))^{p-1}}{\mu^{(k)}((\sigma_i^{(t)}(\mathbf{Z}^{(k+1)}))^{p} + \varepsilon)}, 0\}$.
\ENDFOR\\
\ENDFOR\\
\STATE Update $\mathbf{\Lambda}^{(k+1)}$ and $\mu^{(k+1)}$ by Eq. (\ref{eq: Lagupdate}).
\STATE $k \leftarrow k+1$.
\ENDFOR\\
\ENSURE The completed matrix $\mathbf{X}^{(K)}$.
\end{algorithmic}
\end{algorithm}

\section{Numerical experiments}
In this section, we conduct comprehensive numerical experiments to validate the efficacy of our proposed method. 
To quantitatively assess the reconstruction quality, we employ two widely used metrics: the peak signal-to-noise ratio (PSNR) and the structural similarity index (SSIM) \cite{wang2004image}. 
All experiments were performed in Matlab 2021b with an Intel Core i7-14700KF processor (3.40 GHz), 32 GB of RAM. 
Due to space limitations, more experiments can be found in the Supplemental Material.

\begin{figure}[H]
	\centering
\subfigure{\includegraphics[width=0.075\textwidth]{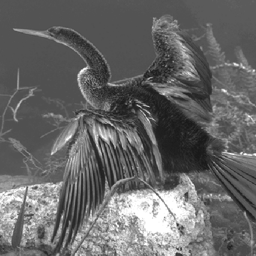}\label{bird}}
\subfigure{\includegraphics[width=0.075\textwidth]{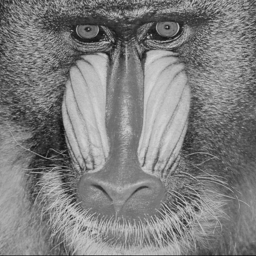}\label{baboon}}
\subfigure{\includegraphics[width=0.075\textwidth]{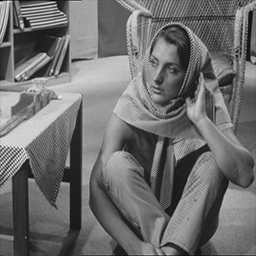}\label{barbara}}
\subfigure{\includegraphics[width=0.075\textwidth]{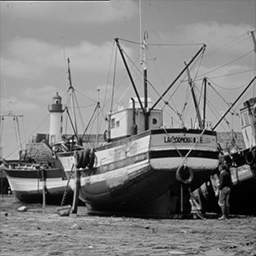}\label{boat}}
\subfigure{\includegraphics[width=0.075\textwidth]{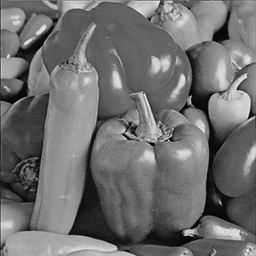}\label{pepper}}
\subfigure{\includegraphics[width=0.075\textwidth]{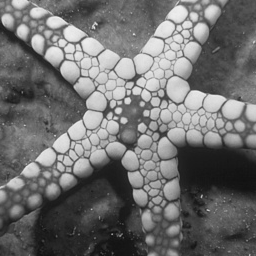}\label{starfish}}
\subfigure{\includegraphics[width=0.075\textwidth]{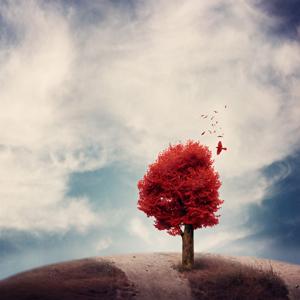}\label{bird}}
\subfigure{\includegraphics[width=0.075\textwidth]{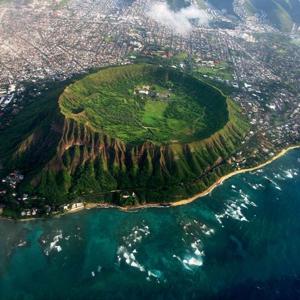}\label{plane}}
\subfigure{\includegraphics[width=0.075\textwidth]{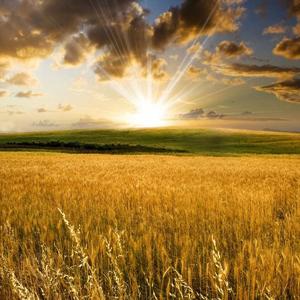}\label{baboon}}
\subfigure{\includegraphics[width=0.075\textwidth]{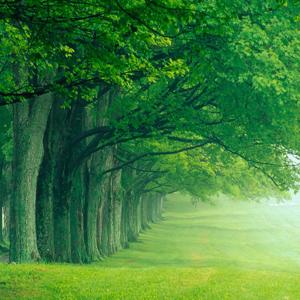}\label{bee}}
\subfigure{\includegraphics[width=0.075\textwidth]{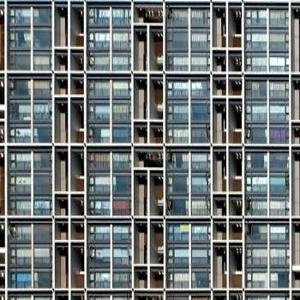}\label{aquatic}}
\subfigure{\includegraphics[width=0.075\textwidth]{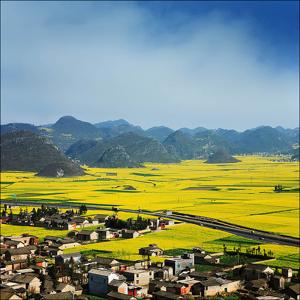}\label{barbara}}
\caption{\small The Set12 dataset. The test images from left-to-right and top-to-bottom are labeled as Img1 to Img12, respectively.}\label{FG:huiOripic}
\end{figure}

\subsection{Experimental Setup}
We compare the proposed RMLN algorithm against several state-of-the-art approaches, categorized as follows:
\begin{itemize}
    \item \textbf{Spectral regularization-based methods:} Geman \cite{kang2015robust}, TNNR (solved via ADMM) \cite{Hu2013TNNM}, WNNM \cite{Gu2017WNNM}, SC$p$ \cite{li2020matrix}, and NMF \cite{shan2023multi}.
    \item \textbf{Matrix factorization-based methods:} D-N \cite{Shang2018DN}, F-N \cite{Shang2018DN}, and LRMF ($n=2$) \cite{chen2021logarithmic}.
\end{itemize}

Our evaluation is conducted on two datasets: Set12 (illustrated in Fig. \ref{FG:huiOripic}) and the benchmark BSD68 \cite{CBSD68}. 
Following prior works \cite{li2020matrix, chen2021logarithmic}, color images are processed individually for each channel.
Our specific parameter settings are as follows: $\lambda=3\times10^5$, $\varepsilon =800$, $\mu^{(0)}=10^{-3}$, $\rho=1.1$, $\gamma=10$, $c= 10^{-8}$, $p=0.8$, $K=100$ and $T=5$. 
To ensure a fair comparison, the parameters for all competing methods were either adopted directly from their original publications or carefully tuned to achieve optimal performance.

\begin{table*}[htbp]
\setlength{\tabcolsep}{10pt}
\renewcommand{\arraystretch}{0.6}
\centering
\caption{\small Average PSNR (dB) and SSIM values of compared methods at different missing ratios (MR = 0.50, 0.65, and 0.75) on the Set12 and BSD68 datasets. Best results are highlighted in \textbf{bold}.}
\begin{center}{ 
\resizebox{\textwidth}{28mm}{
\begin{tabular}{|c||c|c|c|c|c|c|c|c|c|}
\hline
\ Methods & D-N \cite{Shang2018DN} & F-N \cite{Shang2018DN}& Geman \cite{kang2015robust}& TNNR \cite{Hu2013TNNM}& WNNM   \cite{Gu2017WNNM}& SC$p$ \cite{li2020matrix}& NMF \cite{shan2023multi}& LRMF \cite{chen2021logarithmic}& RMLN (ours) \\
\hline
\multicolumn{10}{|c|}{\textbf{Set12}}\\
\hline
\multicolumn{10}{|c|}{MR  = 0.50}\\
\hline
PSNR  & 24.84 & 25.04 & 25.01 & 25.13 & 24.94 & 25.01 & 25.48 & 25.64 & \textbf{26.74}  \\
\hline
SSIM  & 0.7986 & 0.7996 & 0.8029 & 0.8119 & 0.7899 & 0.8007 & 0.8048 & 0.8147 & \textbf{0.8511} \\
\hline

\multicolumn{10}{|c|}{MR  = 0.65}\\
\hline
PSNR   & 22.54 & 22.94 & 22.83 & 22.96 & 22.77 & 22.73 & 23.13 & 23.40 & \textbf{24.24}   \\\hline
SSIM  & 0.7099 & 0.7091 & 0.7143 & 0.7180 & 0.6986 & 0.7061 & 0.7079 & 0.7260 & \textbf{0.7581} \\
\hline

\multicolumn{10}{|c|}{MR  = 0.75}\\
\hline
PSNR  & 20.79 & 21.14 & 21.13 & 21.16 & 20.92 & 21.01 & 21.38 & 21.64 & \textbf{22.40} \\
\hline
SSIM  & 0.6226 & 0.6268 & 0.6417 & 0.6339 & 0.6155 & 0.6269 & 0.6284 & 0.6417 & \textbf{0.6768} \\
\hline

\multicolumn{10}{|c|}{\textbf{BSD68}}\\
\hline
\multicolumn{10}{|c|}{MR  = 0.50}\\
\hline
PSNR  & 24.48 & 25.13 & 25.10 & 25.63 & 24.73  & 24.89 & 25.49 & 25.51 & \textbf{26.80}   \\
\hline
SSIM  & 0.7008 & 0.7215  & 0.7158 & 0.7396 & 0.7021 &  0.7084 & 0.7227  & 0.7272 & \textbf{0.7818} \\
\hline
\multicolumn{10}{|c|}{MR  = 0.65}\\
\hline
PSNR   & 22.96 & 23.31  & 22.95 & 23.05 & 22.64  & 22.69 & 23.25 & 23.24 & \textbf{24.33} \\\hline
SSIM  & 0.6069 & 0.6142  & 0.6026 & 0.6031 & 0.5837 &  0.5943 & 0.6001 & 0.6134 & \textbf{0.6563} \\
\hline

\multicolumn{10}{|c|}{MR  = 0.75}\\
\hline
PSNR   & 21.50 & 21.69  & 21.41 & 21.17 & 21.09 & 20.21 & 21.64 & 21.68 & \textbf{22.59}   \\
\hline
SSIM  & 0.5170 & 0.5101  & 0.5128 & 0.4904 & 0.4847  & 0.5032 & 0.4997 & 0.5225 & \textbf{0.5486} \\
\hline

\end{tabular}}\label{Tb:results1}
}
\end{center}
\end{table*}

\begin{figure*}[]
	\centering

\subfigure[\scriptsize Observed]{\includegraphics[width=0.092\textwidth]{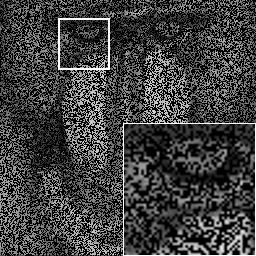}}
\subfigure[\scriptsize D-N \cite{Shang2018DN}]{\includegraphics[width=0.092\textwidth]{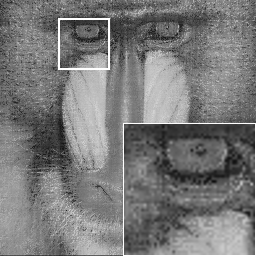}}
\subfigure[\scriptsize F-N  \cite{Shang2018DN}]{\includegraphics[width=0.092\textwidth]{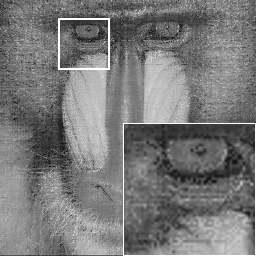}}
\subfigure[\scriptsize Geman \cite{kang2015robust}]{\includegraphics[width=0.092\textwidth]{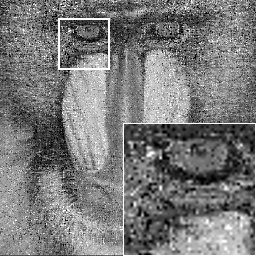}}
\subfigure[\scriptsize TNNR \cite{Hu2013TNNM}]{\includegraphics[width=0.092\textwidth]{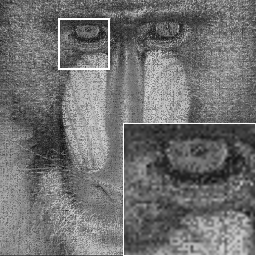}}
\subfigure[\scriptsize WNNM \cite{Gu2017WNNM}]{\includegraphics[width=0.092\textwidth]{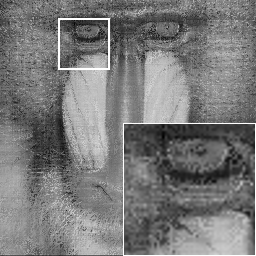}}
\subfigure[\scriptsize SC$p$ \cite{li2020matrix}]{\includegraphics[width=0.092\textwidth]{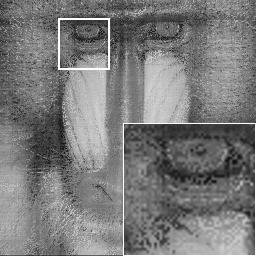}}
\subfigure[\scriptsize NMF \cite{shan2023multi}]{\includegraphics[width=0.092\textwidth]{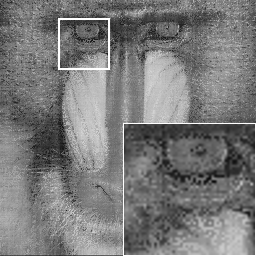}}
\subfigure[\scriptsize LRMF \cite{chen2021logarithmic}]{\includegraphics[width=0.092\textwidth]{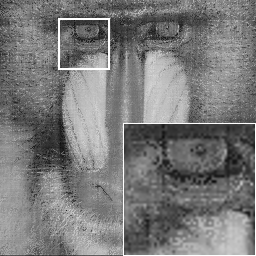}}
\subfigure[\scriptsize RMLN]{\includegraphics[width=0.092\textwidth]{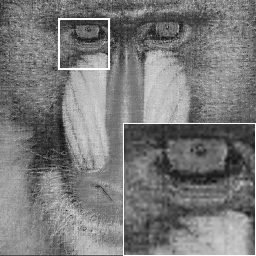}}

\caption{\small The visual quality of different methods on image ``Img2'' from the Set12 dataset with a random mask (MR = 0.50). }\label{FG:showpic1}
\end{figure*}

\begin{figure*}[]
	\centering

\subfigure[\scriptsize Observed]{\includegraphics[width=0.092\textwidth]{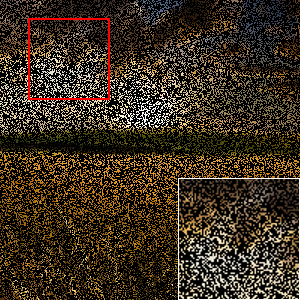}}
\subfigure[\scriptsize D-N \cite{Shang2018DN}]{\includegraphics[width=0.092\textwidth]{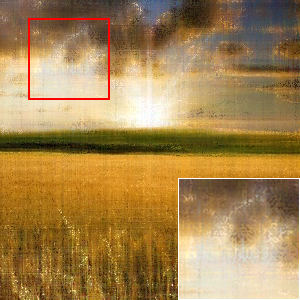}}
\subfigure[\scriptsize F-N \cite{Shang2018DN}]{\includegraphics[width=0.092\textwidth]{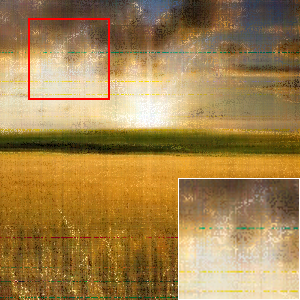}}
\subfigure[\scriptsize Geman \cite{kang2015robust}]{\includegraphics[width=0.092\textwidth]{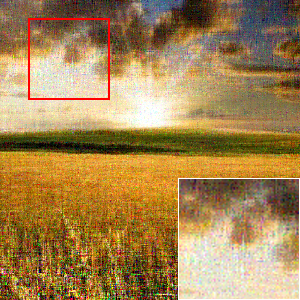}}
\subfigure[\scriptsize TNNR \cite{Hu2013TNNM}]{\includegraphics[width=0.092\textwidth]{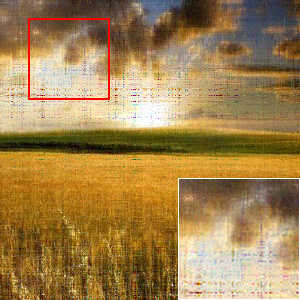}}
\subfigure[\scriptsize WNNM \cite{Gu2017WNNM}]{\includegraphics[width=0.092\textwidth]{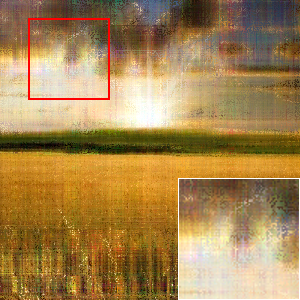}}
\subfigure[\scriptsize SC$p$ \cite{li2020matrix}]{\includegraphics[width=0.092\textwidth]{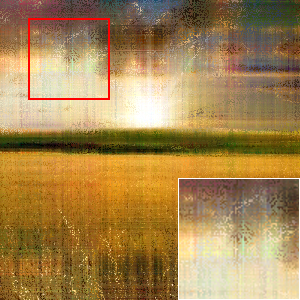}}
\subfigure[\scriptsize NMF \cite{shan2023multi}]{\includegraphics[width=0.092\textwidth]{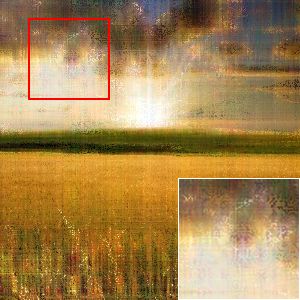}}
\subfigure[\scriptsize LRMF \cite{chen2021logarithmic}]{\includegraphics[width=0.092\textwidth]{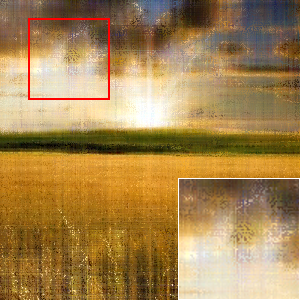}}
\subfigure[\scriptsize RMLN]{\includegraphics[width=0.092\textwidth]{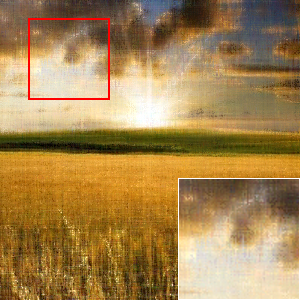}}

\caption{\small The visual quality of different methods on image ``Img9'' from the Set12 dataset with a random mask (MR = 0.65). }\label{FG:showpic2}
\end{figure*}

\subsection{Random Mask Experiments}
We evaluate the performance under random mask scenarios at missing ratios (MRs) of 0.50, 0.65, and 0.75.
Intuitively, a higher MR value corresponds to a more challenging matrix completion task.  
The quantitative results, presented in Table \ref{Tb:results1}, show that our proposed method consistently outperforms other approaches in terms of both PSNR and SSIM across all MR settings. 
For a qualitative assessment, Figs. \ref{FG:showpic1} and \ref{FG:showpic2} provide visual comparisons. 
In contrast to the competing methods, our algorithm produces reconstructions with superior clarity and effectively suppresses undesirable artifacts.

\begin{figure}[]
	\centering
\subfigure[\scriptsize ]{\includegraphics[width=0.063\textwidth]{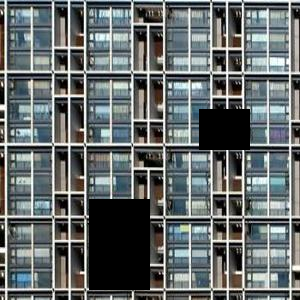}}
\subfigure[\scriptsize ]{\includegraphics[width=0.063\textwidth]{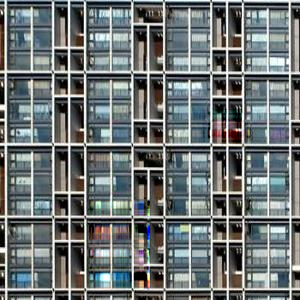}}
\subfigure[\scriptsize ]{\includegraphics[width=0.063\textwidth]{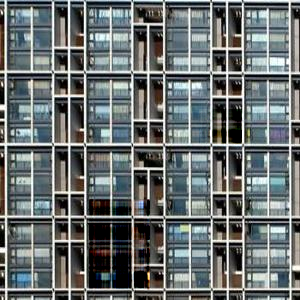}}
\subfigure[\scriptsize ]{\includegraphics[width=0.063\textwidth]{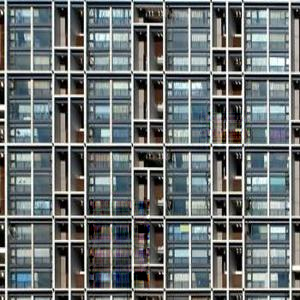}}
\subfigure[\scriptsize ]{\includegraphics[width=0.063\textwidth]{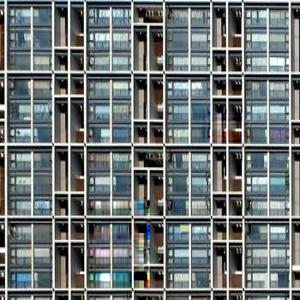}}
\subfigure[\scriptsize ]{\includegraphics[width=0.063\textwidth]{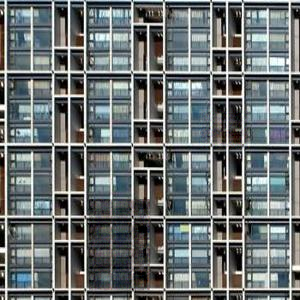}}
\subfigure[\scriptsize ]{\includegraphics[width=0.063\textwidth]{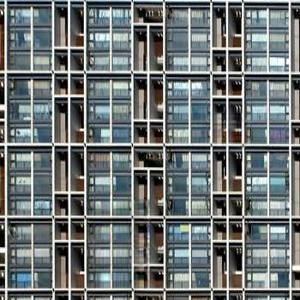}}

\caption{\small The visual quality and SSIM values of different methods on image ``Img11'' from the Set12 dataset with block masks. (a) Observed image; (b) Geman \cite{kang2015robust} (0.9712); (c) TNNR \cite{Hu2013TNNM} (0.9412); (d) WNNM \cite{Gu2017WNNM} (0.9663); (e)  SC$p$ \cite{li2020matrix} (0.9736); (f) NMF \cite{shan2023multi} (0.9658); (g)  RMLN (\textbf{0.9842}). }\label{FG:blockpic2}
\end{figure}

\subsection{Block Mask Experiments}
To further evaluate the generalization capability of our method, we conducted experiments under a more challenging scenario involving block masks. 
Specifically, we applied rectangular occlusion patterns to the test image ``Img11''. 
As illustrated in Fig. \ref{FG:blockpic2}, the visual comparison reveals that existing approaches struggle with this task, either failing to recover structural integrity or producing severe visual artifacts. 
In contrast, our proposed method demonstrates superior performance by effectively preserving fine details and significantly reducing unpleasant artifacts.

\subsection{Role of The Reweighted Strategy}
To investigate the impact of different weighting schemes on the recovery performance, we present a comparative numerical analysis in Table \ref{weihted}. We evaluate three distinct weighting methods: uniform weights $w_i^{\mathbf{a}} = 1$,  logarithmic-based weights  $w_i^{\mathbf{b}} = \gamma\left(\log(\sigma_{i}^p(\mathbf{X}) + \varepsilon) + c \right)^{-1}$ and the reweighted strategy $w_i^{\mathbf{c}} = \gamma\left(\log(\sigma_{i}^p(\mathbf{X}) + \varepsilon) + c \right)^{p-1}$. 
The results in Table \ref{weihted} demonstrate that the reweighted strategy achieves the highest performance metrics.

\begin{table}[]
\centering
\caption{\small Average PSNR (dB) and SSIM values for different weight strategies on the Set12 dataset.}
\small
\begin{center}{ 
\begin{tabular}{|c||c|c|c|}
\hline
\ Weights & MR = 0.50  & MR = 0.65 & MR = 0.75  \\
\hline
\textbf{$w_i^{\mathbf{a}}$}   & 26.03/0.8092 & 23.66/0.7267 & 21.82/0.6390   \\
\hline
\textbf{$w_i^{\mathbf{b}}$}  & 26.59/0.8463 & 24.10/0.7510 & 22.21/0.6673   \\
\hline
\textbf{$w_i^{\mathbf{c}}$}  & 26.74/0.8511 & 24.24/0.7581 & 22.40/0.6768   \\
\hline
\end{tabular}}\
\end{center}\label{weihted}
\end{table}

\begin{figure} 
	\centering
\subfigure[\scriptsize MR =  0.50]{\includegraphics[width=0.14\textwidth]{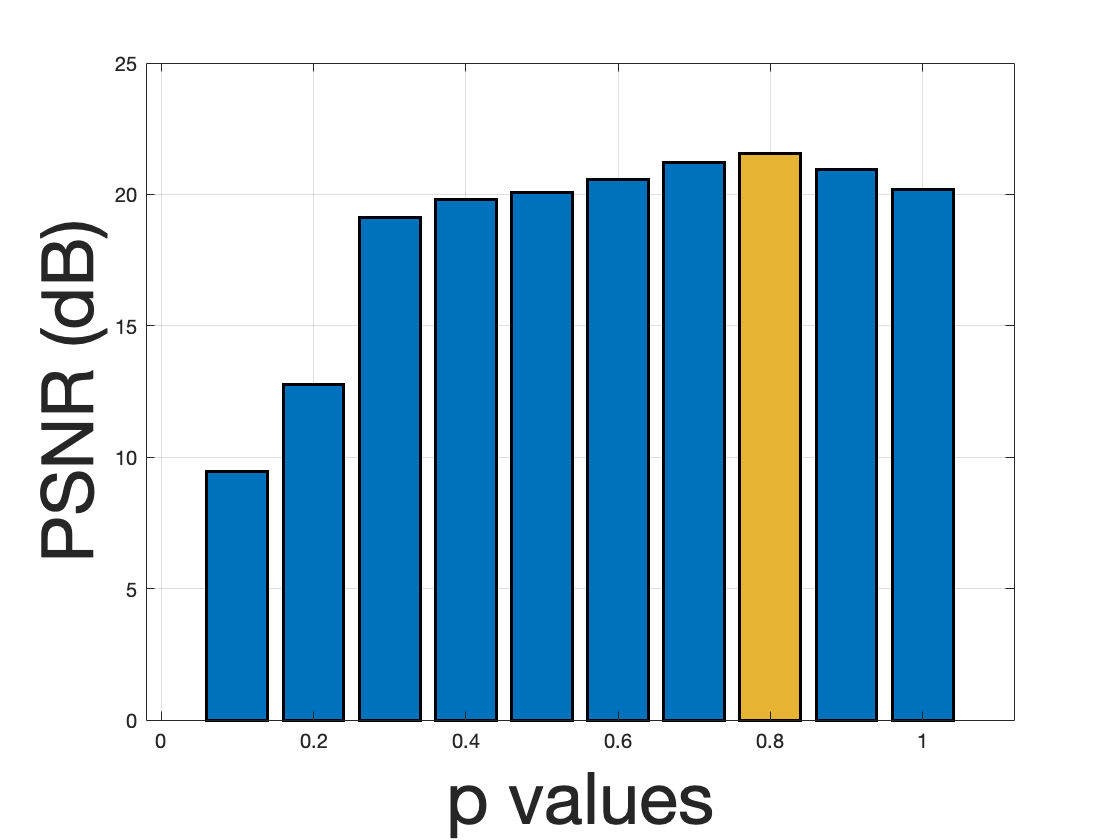}}
\subfigure[\scriptsize MR = 0.65]{\includegraphics[width=0.14\textwidth]{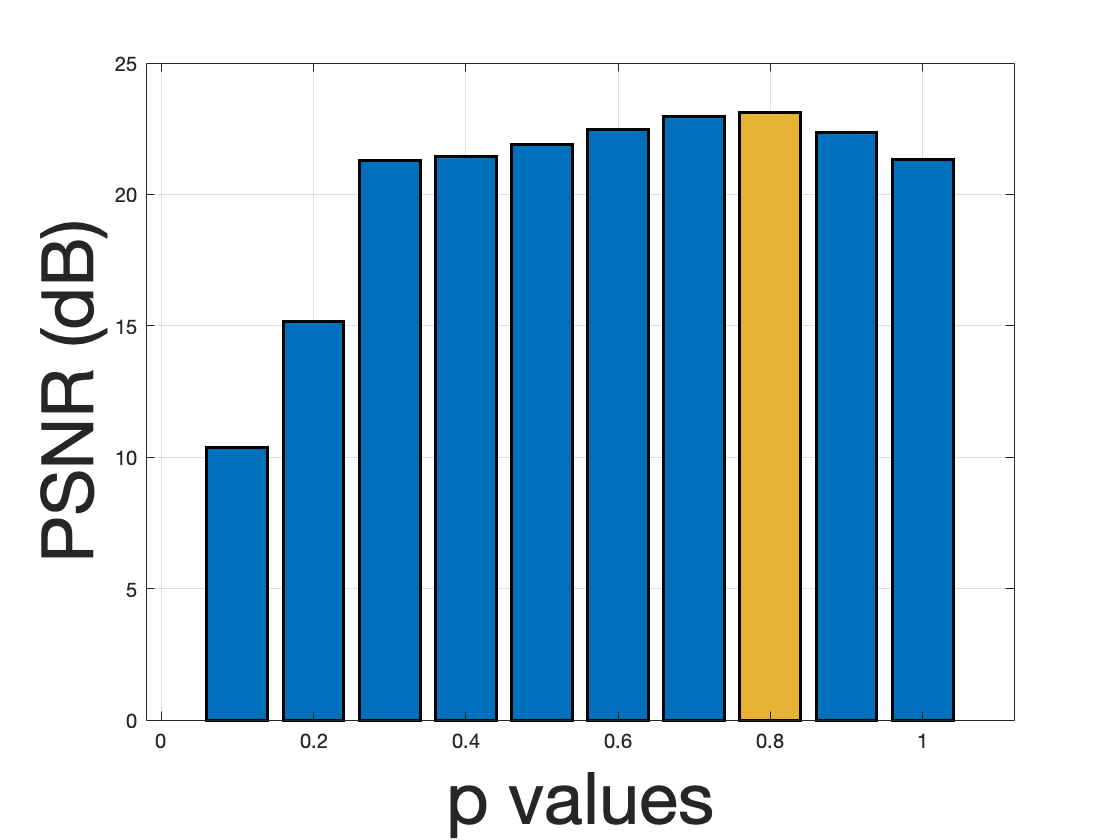}}
\subfigure[\scriptsize MR = 0.75]{\includegraphics[width=0.14\textwidth]{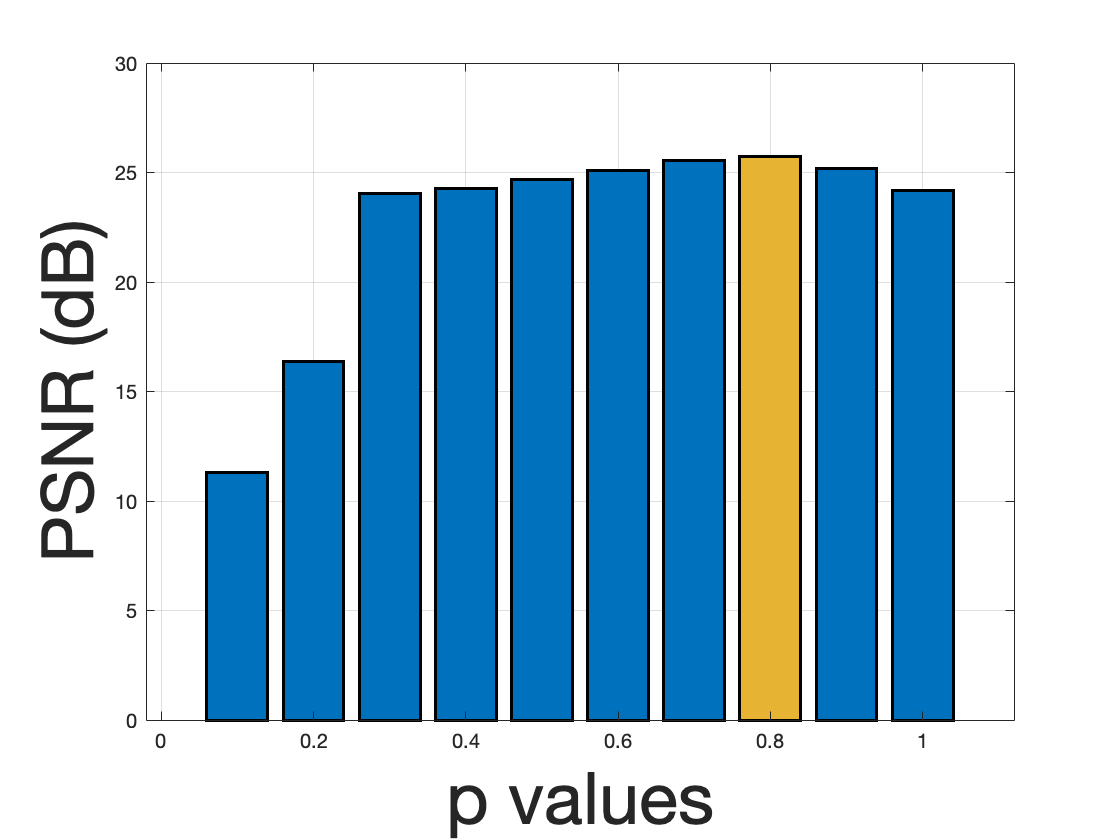}}

\caption{\small The effect of varying $p$ on  ``Img1" under different MRs. }\label{FG:pvalue}
\end{figure}

\subsection{Analysis of Power $p$}
This subsection investigates the impact of the power parameter $p$ within our RMLN minimization framework. 
Fig. \ref{FG:pvalue} illustrates the PSNR values across various MRs as $p$ varies from $0.1$ to $1$. The results indicate that our method is relatively insensitive to the choice of $p$, and empirically, $p=0.8$  is identified as the optimal value.

\section{Conclusion}
This letter introduces a novel reweighted matrix logarithmic norm (RMLN) regularization for matrix completion. The RMLN formulation provides a more accurate approximation to the rank function. Extensive experiments on image inpainting demonstrate that our method achieves superior recovery performance over existing approaches in terms of both quantitative metrics and visual quality. 
Future work will extend the RMLN framework to other relevant tasks, including matrix completion with noise corruption \cite{candes2010matrix,wong2017matrix} and robust principal component analysis \cite{Shang2018DN}.

\bibliography{RLNrefs.bib}%
\bibliographystyle{IEEEtran}

\end{document}